%% file: main.tex
\documentclass{IEEEtran}
\usepackage{graphicx}
\usepackage{amsmath}
\usepackage{cite}
\usepackage{booktabs}
\usepackage[export]{adjustbox}
\usepackage{times}
\usepackage{epsfig}
\usepackage{graphicx}
\usepackage{amsmath}
\usepackage{amssymb}
\usepackage{float}
\usepackage{multirow}
\usepackage{numprint}
\usepackage{wrapfig}
\usepackage[capitalise]{cleveref}

\DeclareRobustCommand\onedot{\futurelet\@let@token\@onedot}
\def\eg{\emph{e.g}. } 
\def\ie{\emph{i.e}. } 
\def\etal{\emph{et al}. } 

\title{Adaptive Voronoi NeRFs}
\author{Tim Elsner, Victor Czech*, Julia Berger*, Zain Selman, Isaak Lim, Leif Kobbelt\\
    Visual Computing Institute, RWTH Aachen University\\
    *equal contribution, \{elsner, zain.selman, isaak.lim\}@cs.rwth-aachen.de}
\begin{document}
\maketitle

\begin{abstract}
Neural Radiance Fields (NeRFs) learn to represent a 3D scene from just a set of registered images. Increasing sizes of a scene demands more complex functions, typically represented by neural networks, to capture all details. Training and inference then involves querying the neural network millions of times per image, which becomes impractically slow. Since such complex functions can be replaced by multiple simpler functions to improve speed, we show that a hierarchy of Voronoi diagrams is a suitable choice to partition the scene. By equipping each Voronoi cell with its own NeRF, our approach is able to quickly learn a scene representation. We propose an intuitive partitioning of the space that increases quality gains during training by distributing information evenly among the networks and avoids artifacts through a top-down adaptive refinement. Our framework is agnostic to the underlying NeRF method and easy to implement, which allows it to be applied to various NeRF variants for improved learning and rendering speeds.
\end{abstract}
\begin{figure}
\begin{center}
\includegraphics[width=0.95\linewidth]{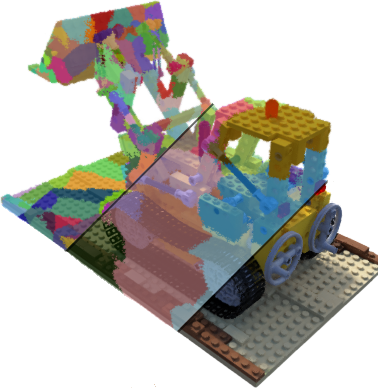}
\end{center}
   \caption{A scene learned with our approach. A Voronoi diagrams subdivides the scene in 16 cells (centre) that are each divided in further 16 cells (left), enabling faster training and inference. The colour of a pixel is always determined by the Voronoi cell of the sample point contributing the most the final colour value.}\label{fig:title}
\end{figure}
\input{1_introduction}
\input{2_related_work}
\input{3_main}
\input{4_evaluation.tex}
\input{5_conclusion.tex}

\bibliographystyle{IEEEtran}
\bibliography{references}

\end{document}

%% file: 1_introduction.tex
\section{Introduction}\label{sec:intro}
Neural Radiance Fields (NeRF)\cite{vanilla} and their derivatives have become one of the most promising areas of research in learning 3D scenes from images. Their key to success lies in using the flexibility of MLPs to learn a volumetric function, guided by the pixel colours of images taken in the scene. For this, pixels are assumed to be the result of integrating a volumetric function over samples from a ray traced through the scene. However, computing even a single pixel requires sampling many times along a ray, \ie many evaluations of the function learning the scene. With larger or more detailed scenes, the complexity of the function must also increase, resulting slower queries for a point. While other approaches often try to improve sampling strategies, our approach takes geometric complexity into account and thus allows for faster convergence and real-time evaluation strategies. Our method can be used complementary to most other techniques and is based on combining three key observations:
\begin{enumerate}
    \item Evaluating a complex function in $k$ places can be much slower than to $k$ times evaluate one of $m$ less complex functions, if choosing the right function to evaluate is cheap enough
    \item Learning a lightweight representation of a scene is a good indicator for what parts of a scene are geometrically complex while having an guidance for the coherence of the scene
    \item Voronoi diagrams can partition a volume into convex cell sections that can be easily evaluated and offer a high degree of flexibility; cells can be easily adjusted to optimise a desired objective
\end{enumerate}
We propose to first learn a simple scene representation that is fast to evaluate for training. This representation is then used to optimise a Voronoi diagram that divides the scene in multiple subsections with roughly the same geometric complexity. The initial simple scene representation then also works as initial representation for the subsections defined by the Voronoi diagram. With this global prior, we are able to avoid coherence problems of approaches that focus on fast evaluation through subdividing a scene and can otherwise be only circumvented with more costly training. This allows us not only to have the speed advantage of distributed scenes for evaluating a learned scene, but to also bring that speed advantage to learning the scene itself.\\
As a result, for a more complex scene, we can use multiple small functions instead of increasing the complexity of one large function. As each sample along a ray through the scene gets evaluated by one function, and we only introduce more low-complexity functions instead of making the function more complex, we can obtain more precise information for a ray sample point without increasing the number of parameters to evaluate for it. This important characteristic is key for large-scale datasets or possibly even real-time approaches in domains such as autonomous driving.\\
As our geometry-aware approach does not alter any of the underlying maths, it is in general compatible with most common variants of NeRFs. While these qualities make our approach particularly more effective with larger amounts of data, it also improves small-scale scenes. It can be implemented as an extension within few lines of simple code, and will be released on GitHub.

%% file: 2_related_work.tex
\section{Related Work}
While learning 3D scene representations from images is a topic with a wide array of classic techniques\cite{tomasi1993shape, hartley2000multiple}, our work builds on techniques that train neural networks to learn a volumetric scene representation. While there are different neural network-based techniques\cite{sitzmann2019scene,lombardi2019neural,mildenhall2019local}, we focus on Neural Radiance Fields.\\

\textbf{Neural Radiance Fields (NeRF)}~~~ as first introduced by Mildenhall \etal \cite{vanilla} learn to produce novel views of a scene from photographs with intrinsic and extrinsic camera parameters. For this, they use a multilayer perceptron (MLP) that should map a position in a scene together with a view direction to a colour and a density value. Treating each pixel as the result of a ray going through the scene, they train the MLP such that volume ray marching on samples of the ray returns the correct pixel colour. Barron \etal \cite{mip} propose mip-NeRF, adapting this formulation to treat pixels as the result of cones going through the scene instead of rays. This better captures volume and helps to combat aliasing effects, synthesising more realistic and sharper views, in particular for datasets that contain images with varying distances to objects. Mip-NeRF 360 \cite{mip360} further improves performance and quality on unbounded scenes by using a non-linear scene parametrisation and by efficiently learning a prior for sampling. Similar scene parametrisations can be found in \cite{donerfdepthoracle, nerf++}. To deal with sparse sets of images, \eg pixelNeRF \cite{pixelnerf} utilizes additional features collected by a CNN. Further work specialises on different aspects:\\
\textbf{Large-Scale Distributed NeRFs}~~~ try to reduce the parameter count of NeRFs for large scenes by breaking the scene into multiple different local NeRFs. This technique enables learning areas spanning multiple hundreds of metres in planar directions \cite{blocknerfwaymo} and can be parallelised during training \cite{meganerf}. Other approaches employ disentanglement to better capture large regions, either by disentangling environment parameters \cite{martin2021nerf}, disentangling the scene itself \cite{zhang2020nerf++, Niemeyer_2021_CVPR}, or by re-parameterising scenes \cite{Park_2021_ICCV, hex}. 

\textbf{Interactive Framerates for NeRFs}~~~ can be achieved by learning a conventional NeRF and then storing the learned opacities and specular features in a sparse voxel grid structure \cite{liu2020neuralsparsevoxelfields, hedman2021snerg}. Thereby, spherical harmonics can be used as a view independent feature representation \cite{yu2021plenoctrees}. While these approaches enable real-time inference rendering by removing costly querying of large networks, they require training a large NeRF beforehand. Rebain \textit{et al.} \cite{derfdecomposed} use a differentiable Voronoi diagram as scene decomposition into many small MLPs to decrease inference time. KiloNeRF \cite{kilonerf} learns a grid of tiny MLPs, enabling real-time rendering. A distillation step from a conventional NeRF is used to avoid artifacts. Kurz \textit{et al.} \cite{kurz2022adanerf} use an efficient sample placement to reduce the number of NeRF evaluations, resulting in real-time capabilities.\\
\textbf{Fast Training of NeRFs}~~~ is a major task in NeRF research, as training can take up to multiple days to reach reasonable output quality. Point-NeRF \cite{pointnerf} generates a point cloud of image features near the surface geometry by using a depth estimation during preprocessing. These local features are then used as prior for fast NeRF training. Other approaches utilise networks trained on tasks like depth prediction to speed up training\cite{deng2022depth, roessle2022dense}. Sun \textit{et al.} \cite{Sun_2022_CVPR} learn a voxel grid of latent features combined with a shallow network to bring down convergence time into the order of minutes. This could be further improved by storing the latent features in a learnable hash table \cite{muller2022instant}.

\textbf{Voronoi Diagrams}~~~ can partition a volume into cells, with points being easily assignable to the according cell. While traditionally used in geometry, optimising Voronoi diagrams can be done \eg to re-create images or shapes\cite{sven, xiao2018optimal, lecot2006ardeco}. Such a partitioning offers a compact and structure-sensitive representation of the underlying information with a high degree of flexibility. The underlying flexibility can also be used to partition a scene for NeRFs, as is done by \cite{derf}.\\
\textbf{Applications}~~~ With widespread use, the applications for NeRF become ever more complex, increasing the demand for faster training. Examples for complex applications are Text-to-NeRF approaches\cite{dreamfusion, jain2022zero, lin2022magic3d} or NeRF-based Text-to-Video approaches\cite{singer2023text}. Other applications include \eg classic tasks from robotics, like localisation and mapping \cite{rosinol2022nerf}. These applications show the need for techniques to consider both speed in training and inference.\\
Our work uses Voronoi diagrams to combine the advantages of approaches that provide interactive framerates at inference together with ideas for large-scale distributed NeRFs.

%% file: 3_main.tex
\begin{figure}[H]
\begin{center}
\includegraphics[width=0.95\linewidth]{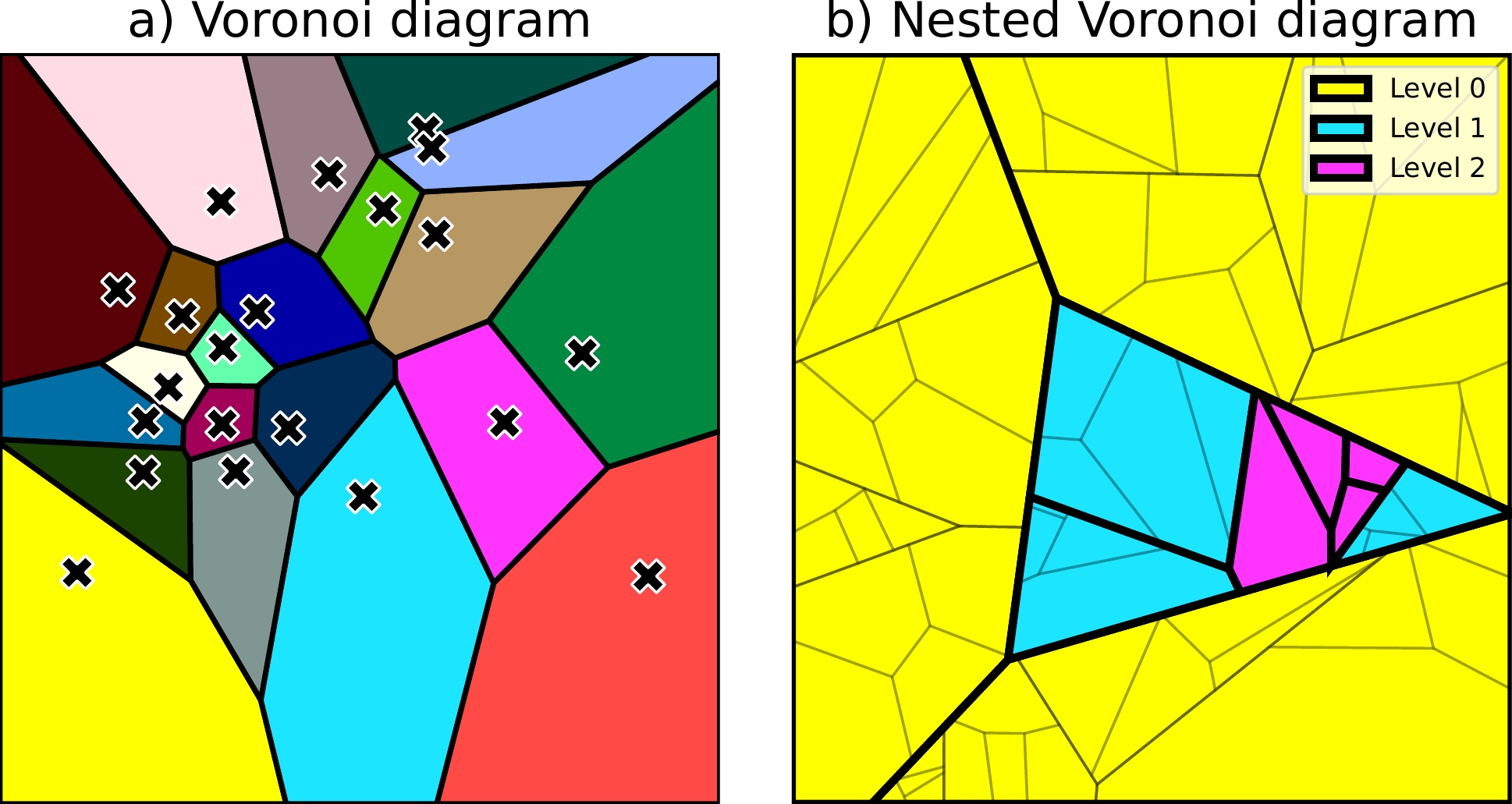}
\end{center}
   \caption{a) An exemplary Voronoi diagram. Each Voronoi cell is displayed in a different colour with its cluster centre as black cross; note the flexibility of the cells covering different amounts of area. b) A nested Voronoi diagram with a depth of two. Each cell is split into 4 smaller cells of the next level.}
\label{fig:voronoi}
\vspace{-0.3cm}
\end{figure}
\begin{figure}[H]
\begin{center}
\includegraphics[width=0.95\linewidth]{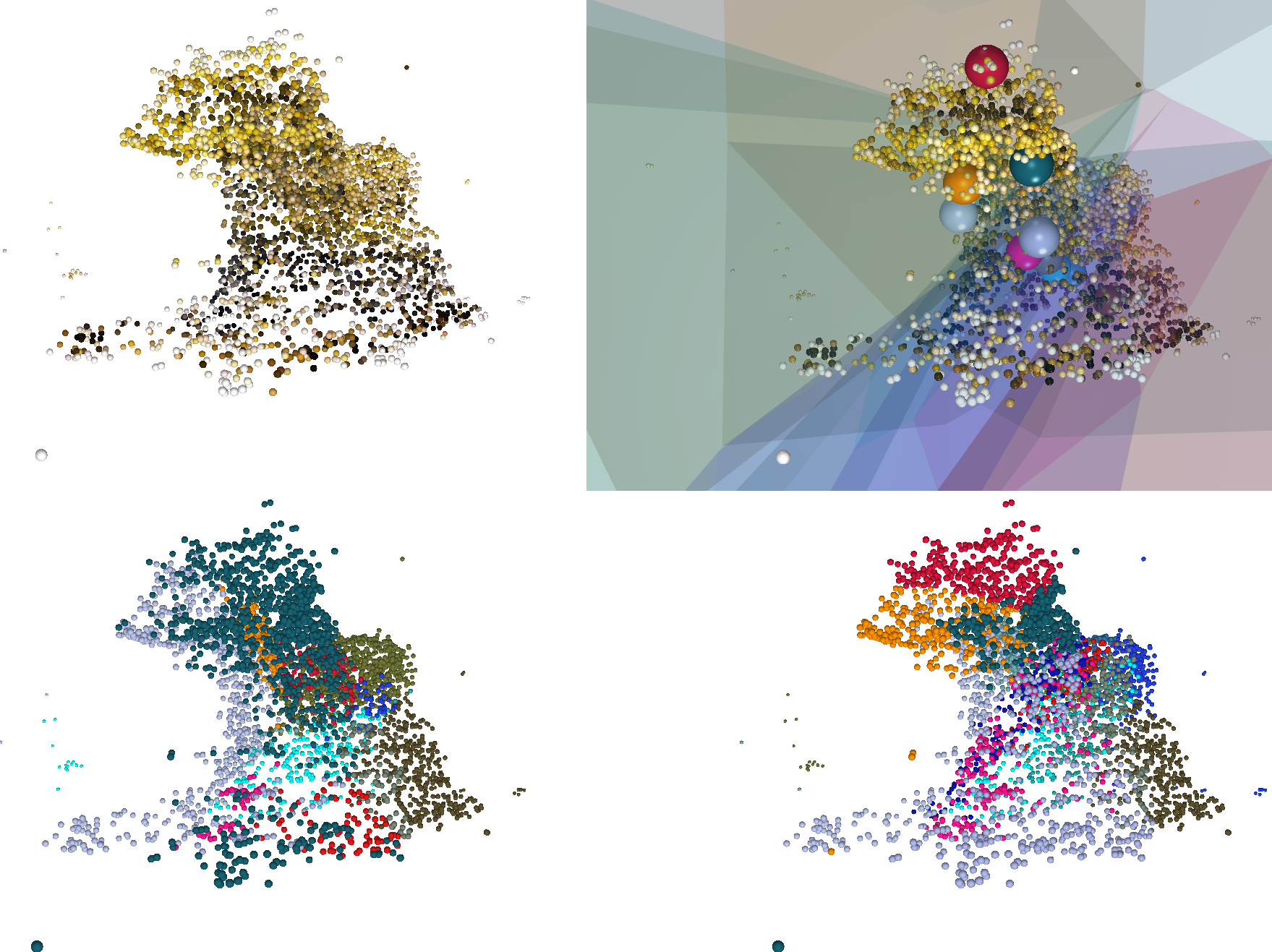}
\end{center}
   \caption{The samples used to optimise the Voronoi cells, visualised as point cloud with their colour values (top left);  point assignments before/after optimisation (bottom left/right), right showing cells more evenly covering the object; the optimised Voronoi diagram with cell centres (top right).}
\label{fig:ptcloud}
\end{figure}

\begin{figure*}
\begin{center}
\includegraphics[width=0.95\linewidth]{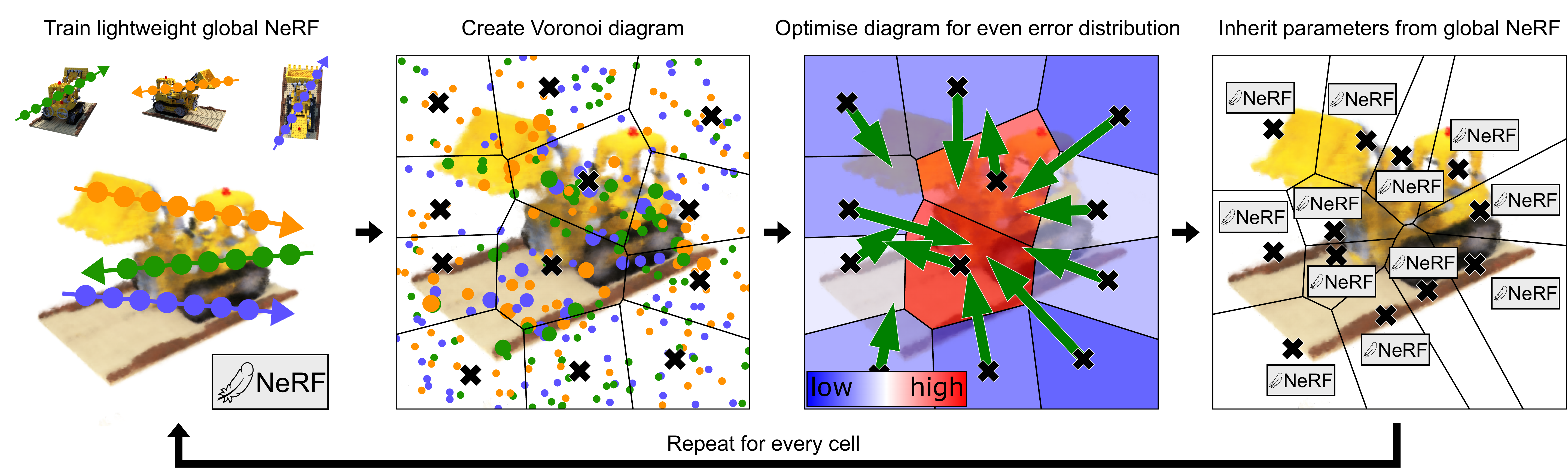}
\end{center}
   \caption{Our proposed approach: A lightweight NeRF learns the global scene (left), then a Voronoi diagram is created to partition the scene space into cells based on weighted ray sample points gathered while training (middle-left). The cells are then optimised with regard to a uniform distribution of error per cell (middle-right). Each cell is given a new lightweight NeRF that inherits the parent parameters. In subsequent training or inference, each ray sample is processed from the according local NeRF of the Voronoi cell it is in (right). The whole pipeline can be applied in a nested fashion.}
\label{fig:architecture}
\end{figure*}

\section{Adaptive Voronoi NeRF}
We introduce Adaptive Voronoi NeRFs, a geometry aware approach that allows faster inference time to the training of distributed NeRFs. By exploiting the geometric information implicitly learned by the NeRF, we achieve better results in training time, inference time, and quality. While distribute the scene among multiple networks, we neither require interpolation between them nor require (possibly not ideal) human-given partitioning of the space.\\
Pixel colour in NeRFs is the result of integrating over samples of a ray going through the volume of a scene where the samples are being evaluated in a neural network that learns to represent a radiance field of the scene. The time consuming procedure of running every ray sample through a large neural network is what our approach accelerates: Instead of learning one large network, we learn many smaller ones, as assigning a point into the according network and querying a smaller network is much faster to evaluate.\\
To understand our idea of accelerating NeRFs with geometric tools, we first re-cap the basics about Voronoi diagrams.
\paragraph{Voronoi Diagrams} are a flexible way to partition a space into convex cells. A Voronoi diagram  $\mathcal{V}$ partitions space by assigning every point to a cell $V_i$ with the closest cell centre $\mathbf{v_i} \in \mathcal{V}$, in our case measured in Euclidean distance. An example can be found in \cref{fig:voronoi}. Formally, for a point $\mathbf{x} \in \mathbb{R}^d$, with $d=3$ in our case, the closest cell centre $\mathbf{v}(\mathbf{x})$ is defined as:
\begin{equation}
\mathbf{v}(\mathbf{x}) = \underset{\mathbf{v_i} \in \mathcal{V}}{\arg \min} ||\mathbf{x}-\mathbf{v_i}||_2
\end{equation}\label{eq:voronoi}
Compared to approaches like octrees or BSP trees, Voronoi diagrams are both flexible in partitioning a space, \eg allowing elongated cells, while also offering cheap point assignment to a cell by simply computing the distance to all cell centres.\\
We can nest Voronoi diagrams by subdividing the content of each cell of an existing diagram into multiple new cells. Assigning a query point into a cell can then be done hierarchically by first finding the correct cell in the first Voronoi diagram, and then finding the correct cell in the Voronoi diagram that subdivides that cell further. This nesting can be done multiple times in a recursive fashion.\\~\\
Based on the three observations made in \cref{sec:intro}, we propose the following approach to adaptively learn a (nested) Voronoi diagram holding a scene (see Figure \ref{fig:architecture} for our visualised pipeline):
\begin{enumerate}
    \item A lightweight NeRF learns the global scene representation for a few iterations
    \item A part of the sample points from regions with dense information are kept, weighted by the error of the ray that are on and the contribution to its final colour
    \item An initially random Voronoi diagram partitions these points and is then optimise to spread the weight evenly, \ie distributing information of the scene evenly among the cells
    \item Each Voronoi cell holds their own new neural network, inheriting the parameters of the global NeRF learned. Training continues with all nets simultaneously, but now evaluates every sample point along a ray with the respective NeRF belonging to the Voronoi cell the point lies in
    \item[] We can apply this adaptive refinement procedure multiple times if necessary, creating new Voronoi diagrams further partitioning cells
\end{enumerate}
This top-down approach is agnostic to technical details of the NeRF, \eg works with considering pixels as either rays\cite{vanilla} or cones\cite{mip} through the volume of the scene, and \eg for approaches with different sampling procedures\cite{kurz2022adanerf}.
\subsection{Estimating Scene Complexity}
We start with training a smaller scene representation, \eg the original NeRF\cite{vanilla}, to learn a rough representation of the entire scene. We use the underlying NeRF architecture, but simply reduce the number of channels in the MLP. To get an estimate of the scene geometry to partition it, we extract a fraction of the most important ray sample points during an epoch, \ie ray samples that contribute the most to the resulting image, for every batch. When only considering one colour channel for simplicity, recall that for a ray $r$, its pixel colour $c_r$, and its \eg 128 samples $s_i, i \in \mathbb{N}_{0}^{<128}$, the resulting loss for updating the NeRF becomes:
\begin{equation}
    E(r) = (c_r - \sum_{i=0}^{127}(w_i \cdot c_i))^2
\end{equation}\label{eq:ray_loss}
Where $w_i$ is the contribution of a sample $s_i$ along the ray to the final pixel colour. As weight to find the most important, \ie heaviest, samples we use a product of the samples contribution to the pixel colour and the error of that pixel, \ie $w_i \cdot E(r)$. While this implicitly takes density of a point into consideration, this also avoids to take ray samples from inside objects, as only the point directly on the surface will contribute much to the rays colour. It also does not overly represent simple, but dense regions, \eg a flat white wall, as these tend to have a low error value. We extract $10000$ ray sample points $S$ per epoch, taking an equal amount of the heaviest points per batch. We then choose $k \in S$, \eg $k=16$, random ray samples as initial Voronoi centres for cells $V_k$.
\subsection{Finding Ideal Voronoi Diagrams}
When formulating a large function as a composition of $k$ smaller functions, making the smaller functions roughly equal in complexity is beneficial: For us, this means equal distribution of scene complexity in the Voronoi diagram cells leads to roughly equal amounts of information to be stored in the respective NeRFs. To optimise the Voronoi centres for this objective, we suggest a simple two step algorithm:\\
First, we assign each sample to their respective Voronoi cell centre, then compute an update for each Voronoi cell centre, and repeat. For the update of the cell centres, we compute update directions to reduce the differences between the total weights $W_i$ of the sample points in each Voronoi cell $V_k$, meaning to even out the amount of information per cell.
For this, we shift light cell centres, \ie cells with little or trivial geometric information in them, towards heavier neighbours, \ie cells with complex geometric information. Likewise, we shift heavy cell away from light centres. For the Voronoi cells, this takes away area from heavy cells and gives it to lighter cells.
As an updated position $\mathbf{v'_i}$ for every cell centre $\mathbf{v_i} \in \mathbb{R}^3$ and its closest 8 Voronoi cells $\mathcal{N}(\mathbf{v_i})$, we compute:
\begin{equation}
    \mathbf{v'_i} = \mathbf{v_i} + \alpha \cdot \sum_{\mathbf{v_j} \in \mathcal{N}(\mathbf{v_i})} \frac{(\mathbf{v_j}-\mathbf{v_i}) \frac{W_j - W_i}{\max_{k}W_k}} {|\mathcal{N}(\mathbf{v_i})|} 
\end{equation}
In result, this computes an update vector for each cell centre by averaging over the weighted directions towards each neighbour. Normalising the weight by dividing through the largest weight in the neighbourhood avoids pushing a cell centre arbitrarily far in a direction. We iterate this optimisation process in parallel for every Voronoi centre, using 500 steps and $\alpha = 0.05$. The assigned positions for each Voronoi cell are not changed anymore after setting them in place. An example of the resulting Voronoi cells can be seen in \cref{fig:ptcloud}.
\subsection{Initialising Cells}
The high quality that NeRFs achieve stems in part from an underlying prior of the network architecture: Multilayer perceptrons (MLPs) learn smooth functions better than noisy ones, and hence are prone to fall into an optimum that is a coherent scene. For multiple MLPs, each MLP in itself has a prior for smoothness, but a combined function of multiple MLPs has no such prior anymore. Hence, it is key to use some sort of prior to ensure scene coherence, as for initialising every cell in a Voronoi diagram with a new, randomly initialised NeRF, produces 'ghosting'-like artifacts as shown in \cref{fig:rnd_vs_mother_img}. In the case of distributed functions that are independent of each other, each function will try to improve the ray's colour even when the object is placed in another cell. While this helps the training objective, it creates complicated, fractured scene representations that perform poorly when evaluating an unseen camera pose (see \cref{fig:rnd_vs_mother_img}).\\
For every new Voronoi cell, we thus initialise the respective NeRF with the parameters of the initially learned, lightweight global scene. Initialising each cell with a NeRF that learned a larger area is giving the cell a prior for shape coherence, hence the optimisation process can generally avoid bad optima. In addition, each cell will converge a bit faster with this initialisation and the boundaries of the cells are already fitting, strongly reducing visible seams even before convergence. Note that the points are always put into the respective NeRF of a cell in global coordinates, as each NeRF inherited the parameters of the NeRF trained with global coordinates.\\
For inference, the only additional burden is the assignment of the correct cell for each sample point, \ie finding the respective NeRF, which is neglectable compared to learning a much larger NeRF. However, optimising multiple NeRFs at once instead of a single NeRF is more costly for backpropagation, but still worth the time saved on the forward pass.\\
We also experimented with a stochastic version of interpolation, taking not always the closest cell centre as the responsible Voronoi cell, but actually sampling the from vector of distances to the cell centres. While this occasionally avoided some small visible seams early in training, it had no more effect after a few epochs of training, as the initial prior from inheriting parameters of a global network was enough.
\subsection{Nested Voronoi Diagrams}
Initially, a learned scene may not cover every area in much detail, hence partitioning a scene into a large number of cells right away can be both impractical and drives up the cost of assigning samples to the right cell. Hence, we propose partitioning every Voronoi cell itself, applying just the same steps as before: Gather all samples that fall into a cell while training, optimise a Voronoi diagram to partition the underlying information evenly, and then give each cell its own neural network that inherits parameters from the parent cell.\\
To decide which cells to subdivide, we have two options: We can either choose to always subdivide the Voronoi cell that is performing worst, \ie accumulates the most error, or simply subdivide every cell at once. As we distribute the cells not only based on density, but also on error, we never experienced cases where one cell in an already partitioned scene was performing significantly worse than another cell. Hence, for simplicity, we subdivide all of the Voronoi cells in parallel.  For all our test scenes, a nesting depth of 2, \ie subdividing the scene once and then subdividing every resulting cell again, was enough. Note that we do not move any existing cell centres once we started training them, as moving them could lead to them covering an area for which they do not have the correct prior.

%% file: 4_evaluation.tex
\begin{figure*}
\begin{minipage}{\linewidth}
  \centering
  \begin{minipage}[t]{0.30\linewidth}
      \begin{figure}[H]
        \begin{center}
        \includegraphics[width=0.95\linewidth]{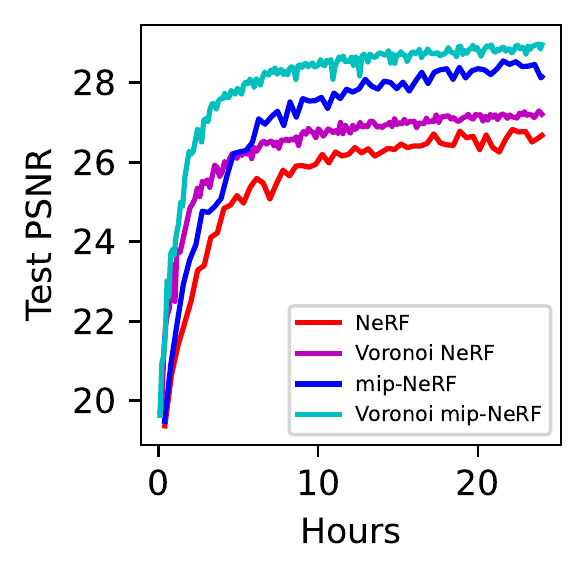}
        \end{center}
           \caption{Test error over time while training either regular or Voronoi variants of NeRF\cite{vanilla} or mip-NeRF\cite{mip} on the ship dataset.}
        \label{fig:single_vs_vornoi}
    \end{figure}
  \end{minipage}
     \hfill
  \begin{minipage}[t]{0.30\linewidth}
      \begin{figure}[H]
        \begin{center}
        \includegraphics[width=1.01\linewidth]{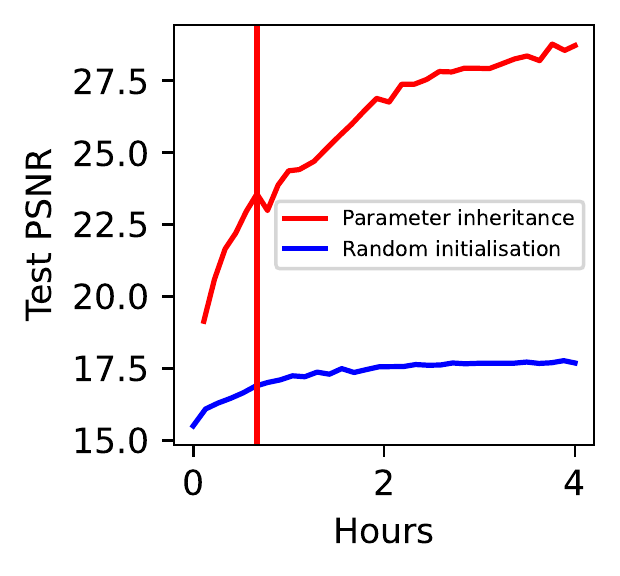}
        \end{center}
           \caption{Test error for cells trained without initialisation (blue), and cells subdivided with copy initialisation (red, subdivision at red line).}
        \label{fig:rnd_vs_mother}
    \end{figure}
  \end{minipage}
     \hfill
  \begin{minipage}[t]{0.30\linewidth}
      \begin{figure}[H]
        \begin{center}
        \includegraphics[width=0.95\linewidth]{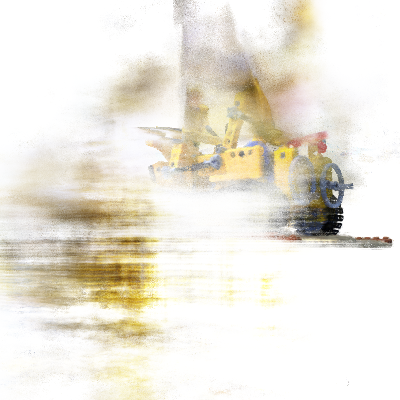}
        \end{center}
           \caption{Artifacts occurring from novel views when not using proper initialisation.}
        \label{fig:rnd_vs_mother_img}
    \end{figure}
  \end{minipage}
\end{minipage}
\end{figure*}
\begin{figure*}
\begin{minipage}{\linewidth}
  \centering
  \begin{minipage}[t]{0.3\linewidth}
      \begin{figure}[H]
        \begin{center}
        \includegraphics[width=0.95\linewidth]{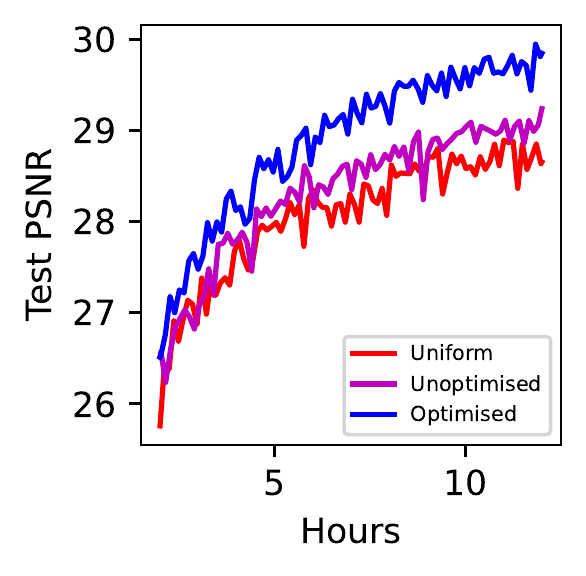}
        \end{center}
           \caption{Comparing uniform cell distribution, random samples as unoptimised cell centres, and optimised cell centres.}
        \label{fig:split_kinds}
    \end{figure}
  \end{minipage}
     \hfill
  \begin{minipage}[t]{0.3\linewidth}
      \begin{figure}[H]
        \begin{center}
        \includegraphics[width=1.01\linewidth]{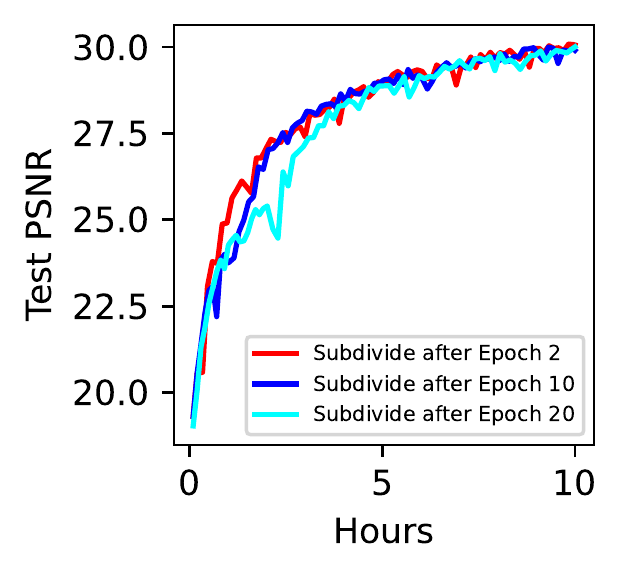}
        \end{center}
           \caption{Early versus late partitioning of the scene. Early can cause tiny artifacts, late can slow learning.}
        \label{fig:earlylate}
    \end{figure}
  \end{minipage}
     \hfill
  \begin{minipage}[t]{0.3\linewidth}
      \begin{figure}[H]
        \begin{center}
        \includegraphics[width=1.01\linewidth]{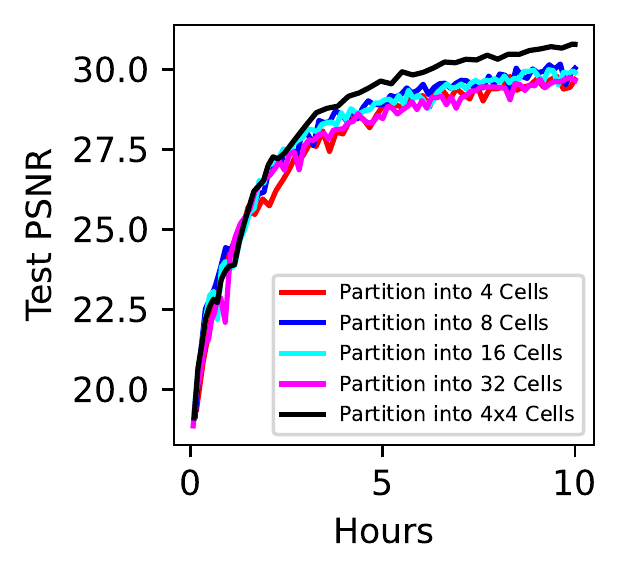}
        \end{center}
           \caption{Different numbers of cells for subdivision, all with roughly the same number of total parameters.}
        \label{fig:cellno}
    \end{figure}
  \end{minipage}
\end{minipage}
\end{figure*}
\begin{table}
    \begin{center}
    \begin{tabular}{|l||c|c|c|}
    \hline
    NeRF & PSNR$\uparrow$ & SSIM$\uparrow$ & LPIP$\downarrow$ \\
    \hline
    Uniform & 25.450 & 0.912 & 0.144\\
    +Voronoi & \textbf{25.603} & \textbf{0.917} & \textbf{0.134}\\
    \hline
    \hline
    mip-NeRF & PSNR$\uparrow$ & SSIM$\uparrow$ & LPIP$\downarrow$ \\
    \hline
    Uniform & 27.401 & 0.945 & 0.111\\
    +Voronoi & \textbf{27.979} & \textbf{0.949} & \textbf{0.105}\\
    \hline
    \end{tabular}
    \end{center}
    \caption{Performance of NeRF\cite{vanilla} and mip-NeRF\cite{mip} for training with stratified sampling instead of hierarchical sampling on the test set after 24 hours of pure training time, compared to their respective Voronoi variants. Bold indicates better.}\label{table:sampling_agnostic}
    \vspace{-0.2cm}
\end{table}
\begin{table}
    \begin{center}
    \begin{tabular}{|l||c|c||c|c|}
    \hline
    PSNR$\uparrow$ & NeRF & +Voronoi & mip-NeRF & +Voronoi \\
    \hline
    Lego & 27.355 & \textbf{27.652} & 30.538 & \textbf{31.234} \\
    Ship & 26.717 & \textbf{27.373} & 28.226 & \textbf{29.007} \\
    Drums & 23.077 & \textbf{23.490} & 24.474 & \textbf{24.760} \\
    \hline
    SSIM$\uparrow$ & NeRF & +Voronoi & mip-NeRF & +Voronoi \\
    \hline
    Lego & 0.9451 & \textbf{0.9483} & 0.9712 & \textbf{0.9752} \\
    Ship & 0.8967 & \textbf{0.9068} & 0.9164 & \textbf{0.9250} \\
    Drums & 0.9145 & \textbf{0.9230} & 0.9354 & \textbf{0.9409} \\
    \hline
    LPIP$\downarrow$ & NeRF & +Voronoi & mip-NeRF & +Voronoi \\
    \hline
    Lego & 0.0910 & \textbf{0.0831} & 0.0507 & \textbf{0.0417} \\
    Ship & 0.2016 & \textbf{0.1820} & 0.1658 & \textbf{0.1439} \\
    Drums & 0.1320 & \textbf{0.1128} & 0.1054 & \textbf{0.0869} \\
    \hline
    \end{tabular}
    \end{center}
    \caption{Performance of NeRF\cite{vanilla} and mip-NeRF\cite{mip} on the test set of different datasets using hierarchical sampling after 24 hours of pure training time, compared to their respective Voronoi variants. Bold indicates better. See \cref{fig:single_vs_vornoi} for convergence behavior.}\label{table:architecture_agnostic}
    \vspace{-0.5cm}
\end{table}
\section{Evaluation}
Our proposed method is built to be an independent extension for existing methods that speeds up train and inference time without sacrificing quality or causing artifacts.\\
We first evaluate possible hyper parameter choices, then provide an ablation to highlight the impact of all our components (\cref{sec:eval_ablation}). We then provide experiments and argumentation to why our approach is agnostic to \eg the underlying foundation for sampling strategies, underlying NeRF architecture, and ray formulation (\cref{sec:eval_agnostic}). Then, we discuss our approach in comparison to other work in terms of evaluation speed, quality, and benefit for large-scale scenes (\cref{sec:comparison}).\\
Throughout this evaluation, we focus our evaluation on train speed, accuracy on test images, and inference speed. For fairness, we always measure performance over training time, as measuring with epochs alone would not take the computational overhead of our method, \ie assigning points into their cells, into consideration. We would also gain an unfair advantage from our much shorter inference times for the test set. If not specified otherwise, due to limited resources, our experiments are run on 400-by-400 pixel versions of sets, if not further specified the Lego model, from the NeRF datasets\cite{vanilla} with a single GeForce RTX 2080 Ti. We use 256 channels (one single network) versus 64 channels (Voronoi variants) resulting in 12 cells with slightly fewer total parameters for the different NeRF architectures, and train for 24 hours. Note that we tested our approach with simple PyTorch code, with no refined performance boosts or possibly even specific CUDA optimisations, for better comparability. We limit ourselves to a single level of subdivision for simplicity, while we did observe best performance when using multiple levels (see \cref{fig:cellno}). We always measure the PSNR of the mean squared error on the whole testset for any plots and otherwise give the average PSNR per image over the testset. In general, we show that our approach improves convergence speed and quality, as can be seen in \cref{fig:single_vs_vornoi}.\\
\textbf{Limitations}~~~ While our approach is only an extension, hence inherits weaknesses from the underlying NeRFs, our approach has an additional weakness shown by short training time for the global scene: Our approach struggles with scenes with initially unclear geometry, \eg many translucent objects, as the initially learned scene will struggle to provide a meaningful geometric prior and may lead to a bad subdivision that can not be changed anymore over the training.
\subsection{Hyper Parameter Choices and Ablation}\label{sec:eval_ablation}\label{sec:eval_hyper}
\textbf{Subdivision strategies}~~~Optimising Voronoi cells is key to the success of our approach: We want to distribute information evenly between the cells to make best use of each network for fast convergence and high quality. Large differences between the total weight of our cells would indicate a bad distribution, and similar total weights would indicate a good distribution. Our proposed algorithm reduces the median/average standard deviation over 100 examples on different datasets from $~0.2458 / ~0.2479$ to $~0.0018 / 0.2479$. In consequence, networks using Voronoi diagrams with optimised cell positions show a clear advantage, as shown in \cref{fig:split_kinds}.\\
\textbf{Number of cells}~~~For the decision on the number of cells, we evaluate 4, 8, 16, and 32 cells, while also considering 4 cells subdivided in 4 further cells (16 in total), each with roughly the same number of total parameters. As can be seen in \cref{fig:earlylate}, subdividing early is no issue, but can cause tiny artifacts not impacting the scores, while subdividing too late wastes computation time. Similarly, \cref{fig:cellno} shows that too few or too many cells slow convergence.\\
\textbf{Initialisation}~~~Training a distributed network without the right prior can lead to failure of generalisation. The result of this are local optima that prevent generalisation and that can cause artifacts from novel views (see \cref{fig:rnd_vs_mother} and \cref{fig:rnd_vs_mother_img}). These artifacts are results of improving train quality by creating dense regions where there should not be any. Inheriting network parameters from a cell with larger global context prevents this effectively, while not requiring any form of distillation as in \eg KiloNeRF\cite{kilonerf}. We also observe no issues with the learned geometric representation that this prior affects, even at times better learning the scene geometry, as can be seen from the visualisation of the total density for each ray, see inset (left: mip-NeRF, right: Voronoi mip-NeRF). \begin{wrapfigure}{l}{0.25\textwidth}
    \vspace{-0.4cm}
    \centering
    \includegraphics[width=0.9\linewidth]{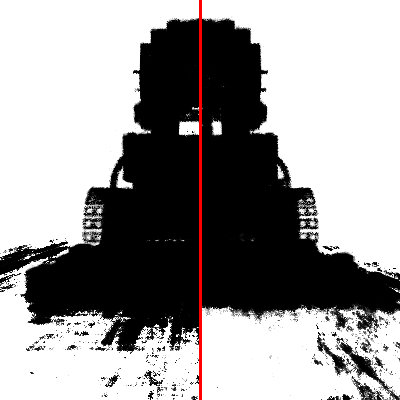}
    \label{fig:ray_density}
    \vspace{-1.cm}
\end{wrapfigure}
\subsection{Independence of Underlying Architecture}\label{sec:eval_agnostic}
Our approach introduces a geometrically inspired dynamically learned decomposition of the NeRF problem into many localised, smaller problems. Since the underlying mathematical formulation is not altered, this speedup from decomposition can be easily combined with many other improvements introduced subsequently to the initial NeRF.
We demonstrate the effectiveness of our approach on different NeRF variants by investigating different sampling strategies, different approaches to the ray formulation, and by exploring further approaches that have recently gathered attention.\\
First, we show that our Voronoi NeRF approach is improving results regardless of the sampling strategies used. For this, we compare results of stratified and hierarchical sampling with either mip-NeRF \cite{mip} and \cite{vanilla}, showing clear improvement for both stratified (see \cref{table:sampling_agnostic}) and hierarchical samplings strategies (see \cref{table:architecture_agnostic}). In more detail, we then demonstrate that Voronoi NeRF as an extension for different architectures, namely NeRF\cite{vanilla} or mip-NeRF\cite{mip}, outperforms their counterparts on various datasets, see \cref{table:architecture_agnostic}. We also show that our approach can work with the small network architecture proposed by KiloNeRF\cite{kilonerf} in \cref{sec:comparison}.\\
In summary, we provide improvements in particular early in training, converging much faster while having fewer fluctuations during training, as can be seen in \cref{fig:single_vs_vornoi}.
We further discuss how our approach is agnostic to other architectures and approaches: mip-NeRF 360 \cite{mip360} and DONeRF \cite{donerfdepthoracle} use a non-linear contraction function to restrict the coordinate range of points far from the origin. Furthermore, additional simplified NeRF-like networks are often used for more feasible samplings \cite{vanilla}\cite{mip360}. These extensions are in no way in conflict with our approach, in fact one might even accelerate the sample predictor networks by localisation as well. Inference accelerations such as precomputed opacity grids, empty space skipping and early ray termination \cite{kilonerf} are also complementary to our approach, as they do not change the actual NeRF mechanism. Interpolating outputs of multiple cells can be used to smooth transitions of neighboring cells \cite{blocknerfwaymo} or improve cell placement \cite{derfdecomposed}. While we explored this and saw improvements in early stages about occasionally visible, very tiny seams, the prior given through initialisation with a global NeRF is enough to avoid any visible seams after the first few epochs after a subdivision.
\subsection{Comparison to Others}\label{sec:comparison}
As discussed in \cref{sec:eval_agnostic}, our approach is compatible with many improvements suggested for NeRFs when it comes to training speed and inference quality. However, the faster training times obtained through an approach exploiting geometric knowledge can also be used for both faster inference and learning larger scenes faster and in better quality. KiloNeRF\cite{kilonerf} provides faster inference by first training a large NeRF that is then distilled into many smaller NeRFs arranged in a grid fashion. They obtain further speed by optimising the sampling process, \eg skipping largely empty sections, terminating rays early, and optimising CUDA code. As all of these options are available to our approach as well, we only discuss the qualitative results. Analogue to their distillation, our parameter inheritance works in a top-down fashion while training the network, avoiding the extra step of distilling. For comparison and as another example for the flexibility towards the underlying architecture, we trained a Voronoi NeRF with 16 cells, subdivided each cell into another 16 cells, obtaining 256 cells in total. We compare to their approach with 512 uniformly distributed cells, trained for the same duration. We use the same architecture as they do, and use our inheritance initialisation instead of their distillation. Effectively, this can be considered an on-the-fly distillation process from coarse to fine. With the same inference speed and no need for the extra distillation step, and geometry-sensitive cell distribution, our approach can outperform KiloNeRF in terms of quality at half the number of cells, see \cref{table:vs_kilonerf}. We attribute this to our representation making better use of its parameters by making sure every cell is filled with about the same amount of information, where KiloNeRF can place cells entirely inside an object or in thin air.\\
As can be seen in \cref{fig:single_vs_vornoi}, our approach trains faster later on through multiple smaller networks, while particularly the early convergence benefits from having only one small network that learns the scene. Our top-down parameter inheritance can thus boost performance in particular early for scenes that are far from convergence. These two qualities make our approach particularly valuable for large datasets like the ones proposed by MegaNeRF\cite{meganerf} or Block-NeRF\cite{blocknerfwaymo}, as the larger or more detailed the scene grows, only our number of networks increases while individual ray sample evaluations, except for the relatively cheap assignment of each sample to a cell, do not become any more expensive. We further argue that, as we have shown, dynamically partitioned space leads to better performance. In particular, our results indicate that adaptively subdividing a scene more than twice will bring even more relative improvement than for the tested small-scale scenes.
\input{4_figures_eval.tex}
In summary, our approach does significantly speed up training, allows for fast inference, and is invariant to the underlying architecture and other degrees of freedom within the NeRF formulation. Opposing to other distributed approaches, it does not require any expensive precomputation\cite{kilonerf}, requires no interpolation between cells\cite{meganerf,blocknerfwaymo,derf}, and does not require previous knowledge of the scene, \eg as human-given specification for the layout of the learned partition\cite{meganerf,blocknerfwaymo,kilonerf}.

%% file: 4_figures_eval.tex
\begin{table}
    \begin{center}
    \begin{tabular}{|l | c | c | c|}
    \hline
      Approach & PSNR$\uparrow$ & SSIM$\uparrow$ & LPIP$\downarrow$ \\ \hline
      KiloNeRF & 27.080 & 0.938 & 0.098 \\ \hline
      Voronoi & \textbf{27.765} & \textbf{0.952} & \textbf{0.078} \\
    \hline
    \end{tabular}
    \end{center}
    \caption{Quality of KiloNeRF\cite{kilonerf} compared to our Voronoi version with the same architecture and speed after the same amount of training.}\label{table:vs_kilonerf}
\end{table}

%% file: 5_conclusion.tex
\section{Conclusion}
We propose an easy to use extension for Neural Radiance Fields that allows us to bring the quicker inference times from distributed approaches to training. We achieve this through considering the scene to be a (nested) Voronoi diagram that is adaptively refined through the training process. We build this diagram through exploiting the geometric information learned while training and reduce artifacts by obtaining a prior for a coherent shape from passing down parameters from a global to a local scale. Our approach achieves high speed from subdivision of the scene into networks that are fast to query, achieves high quality from geometry-sensitive adaptive space partitioning, and uses inheritance initialisation to avoid artifacts. As this solution is agnostic to architecture, sampling, and even conceptual differences in aspects like considering rays or cones, our approach works with many different NeRF variants. It can improve approaches that are built for speed and push the qualitative boundaries for existing approaches. With its dynamic adaptivity in refining detail, it offers flexibility and speed for \eg large datasets, an area where NeRFs usually require costly hardware and hand-tailored solutions. With our approach, we provide a simple extension to bring the field closer to the masses on a hardware level, while its applicability to many different kinds of NeRF approaches and its simplicity make it accessible not only to experts in the field.
For future work, we see possibilities to use our geometry prior even more adaptively: For dynamic scenes with an initialisation, these would become easy to adapt, as only the Voronoi cell(s) with the ongoing change would need to be updated. We also see that our approach could be used for formulations that learn \eg signed distance functions.